\documentclass[11pt,a4paper]{article}
\usepackage[hyperref]{acl2018}
\usepackage{times}
\usepackage{latexsym}
\usepackage{xcolor}
\usepackage{multirow}
\usepackage{graphicx}
\usepackage{subfig}

\usepackage{url}

\aclfinalcopy 

\title{Findings of the Second Workshop on \\ Neural Machine Translation and Generation}

\author{Alexandra Birch$^{\spadesuit}$, Andrew Finch$^{\heartsuit}$, Minh-Thang Luong$^{\clubsuit}$, Graham Neubig$^{\diamondsuit}$, Yusuke Oda$^{\circ}$\\
  $^\spadesuit$University of Edinburgh,
  $^\heartsuit$Apple,
  $^\clubsuit$Google Brain,
  $^\diamondsuit$Carnegie Mellon University
  $^\circ$Google Translate}

\date{}

\begin{document}
\maketitle
\begin{abstract}
  This document describes the findings of the Second Workshop on Neural Machine Translation and Generation, held in concert with the annual conference of the Association for Computational Linguistics (ACL 2018).
  First, we summarize the research trends of papers presented in the proceedings, and note that there is particular interest in linguistic structure, domain adaptation, data augmentation, handling inadequate resources, and analysis of models.
  Second, we describe the results of the workshop's shared task on efficient neural machine translation (NMT), where participants were tasked with creating NMT systems that are both accurate and efficient.
\end{abstract}

\section{Introduction}
\label{sec:intro}

Neural sequence to sequence models \cite{kalchbrenner13rnntm,sutskever14sequencetosequence,bahdanau15alignandtranslate} are now a workhorse behind a wide variety of different natural language processing tasks such as machine translation, generation, summarization and simplification.
The 2nd Workshop on Neural Machine Translation and Generation (WNMT 2018) provided a forum for research in applications of neural models to machine translation and other language generation tasks (including summarization \cite{rush2015neuralattention}, NLG from structured data \cite{wen2015conditioned}, dialog response generation \cite{vinyals2015neural}, among others).
Overall, the workshop was held with two goals:

First, it aimed to synthesize the current state of knowledge in neural machine translation and generation: This year we will continue to encourage submissions that not only advance the state of the art through algorithmic advances, but also analyze and understand the current state of the art, pointing to future research directions.
Towards this goal, we received a number of high-quality research contributions on the topics of linguistic structure, domain adaptation, data augmentation, handling inadequate resources, and analysis of models, which are summarized in Section \ref{sec:researchprogram}.

Second, it aimed to expand the research horizons in NMT: Based on panel discussions from the first workshop, we organized a shared task.
Specifically, the shared task was on ``Efficient NMT''. The aim of this task was to focus on not only accuracy, but also memory and computational efficiency, which are paramount concerns in practical deployment settings.
The workshop provided a set of baselines for the task, and elicited contributions to help push forward the Pareto frontier of both efficiency and accuracy.
The results of the shared task are summarized in Section \ref{sec:sharedtask}

\section{Summary of Research Contributions}
\label{sec:researchprogram}

We published a call for long papers, extended abstracts for preliminary work, and cross-submissions of papers submitted to other venues. The goal was to encourage discussion and interaction with researchers from related areas. 
We received a total of 25 submissions, out of which 16 submissions were accepted. The acceptance rate was 64\%. Three extended abstracts, two cross-submissions and eleven long papers were accepted after a process of double blind reviewing. 

Most of the papers looked at the application of machine translation, but there is one paper on abstractive summarization~\citep{fan18abssum} and one paper on automatic post-editing of translations~\citep{unanue18postedit}. 

The workshop proceedings cover a wide range of phenomena relevant to sequence to sequence model research, with the contributions being concentrated on the following topics: 
\begin{description}
\item[Linguistic structure:] How can we incorporate linguistic structure in neural MT or generation models? Contributions examined the effect of considering semantic role structure~\citep{marcheggiani18gcn}, latent structure~\citep{bastings18latentstruct}, and structured self-attention~\citep{bisk18grammar}.
\item[Domain adaptation:] Some contributions examined regularization methods for adaptation~\citep{khayrallah18regdom} and ``extreme adaptation'' to individual speakers~\citep{michel18extremenmt}
\item[Data augmentation:] A number of the contributed papers examined ways to augment data for more efficient training. These include methods for considering multiple back translations~\citep{imamura18selftraining}, iterative back translation~\citep{hoang18backtrans}, bidirectional multilingual training~\citep{niu18bidirect}, and document level adaptation~\citep{kothur18docadapt}
\item[Inadequate resources:] Several contributions involved settings in which resources were insufficient, such as investigating the impact of noise~\citep{khayrallah18noise}, missing data in multi-source settings~\citep{nishimura18missingdata} and one-shot learning~\citep{pham18oneshot}.
\item[Model analysis:] There were also many methods that analyzed modeling and design decisions, including investigations of individual neuron contributions~\cite{bau18neurons}, parameter sharing  ~\citep{jean18sharing}, controlling output characteristics~\citep{fan18abssum}, and shared attention~\cite{unanue18postedit}
\end{description}


\section{Shared Task}
\label{sec:sharedtask}

Many shared tasks, such as the ones run by the Conference on Machine Translation \cite{bojar2017wmt}, aim to improve the state of the art for MT with respect to accuracy: finding the most accurate MT system regardless of computational cost.
However, in production settings, the efficiency of the implementation is also extremely important.
The shared task for WNMT (inspired by the ``small NMT'' task at the Workshop on Asian Translation \cite{nakazawa2017wat}) was focused on creating systems for NMT that are not only accurate, but also efficient.
Efficiency can include a number of concepts, including memory efficiency and computational efficiency.
This task concerns itself with both, and we cover the detail of the evaluation below.

\subsection{Evaluation Measures}
\label{sec:eval}

The first step to the evaluation was deciding \emph{what} we want to measure.
In the case of the shared task, we used metrics to measure several different aspects connected to how good the system is.
These were measured for systems that were run on CPU, and also systems that were run on GPU.

\begin{description}
\item[Accuracy Measures:] As a measure of translation accuracy, we used BLEU \cite{papineni02bleu} and NIST \cite{doddington02nist} scores.
\item[Computational Efficiency Measures:] We measured the amount of time it takes to translate the entirety of the test set on CPU or GPU. Time for loading models was measured by having the model translate an empty file, then subtracting this from the total time to translate the test set file.
\item[Memory Efficiency Measures:] We measured: (1) the size on disk of the model, (2) the number of parameters in the model, and (3) the peak consumption of the host memory and GPU memory.
\end{description}

These metrics were measured by having participants submit a container for the virtualization environment Docker\footnote{\url{https://www.docker.com/}}, then measuring from outside the container the usage of computation time and memory.
All evaluations were performed on dedicated instances on Amazon Web Services\footnote{\url{https://aws.amazon.com/}}, specifically of type \texttt{m5.large} for CPU evaluation, and \texttt{p3.2xlarge} (with a NVIDIA Tesla V100 GPU).

\subsection{Data}

The data used was from the WMT 2014 English-German task \cite{bojar14wmt}, using the preprocessed corpus provided by the Stanford NLP Group\footnote{\url{https://nlp.stanford.edu/projects/nmt/}}.
Use of other data was prohibited.

\begin{figure*}[t!]
  \centering
  \subfloat[CPU Time vs. Accuracy]{\includegraphics[width=0.49\textwidth]{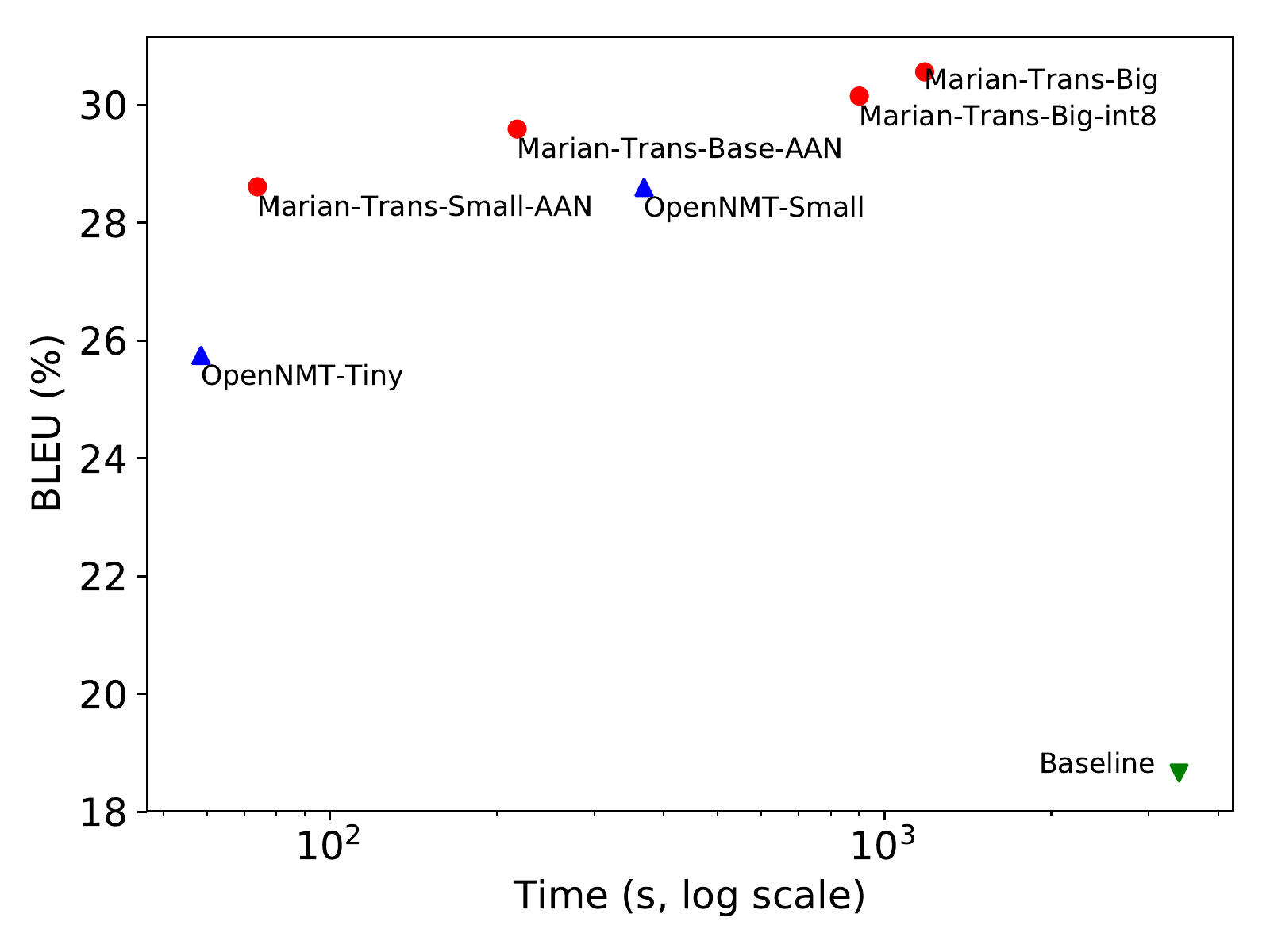} }
  \subfloat[GPU Time vs. Accuracy]{\includegraphics[width=0.49\textwidth]{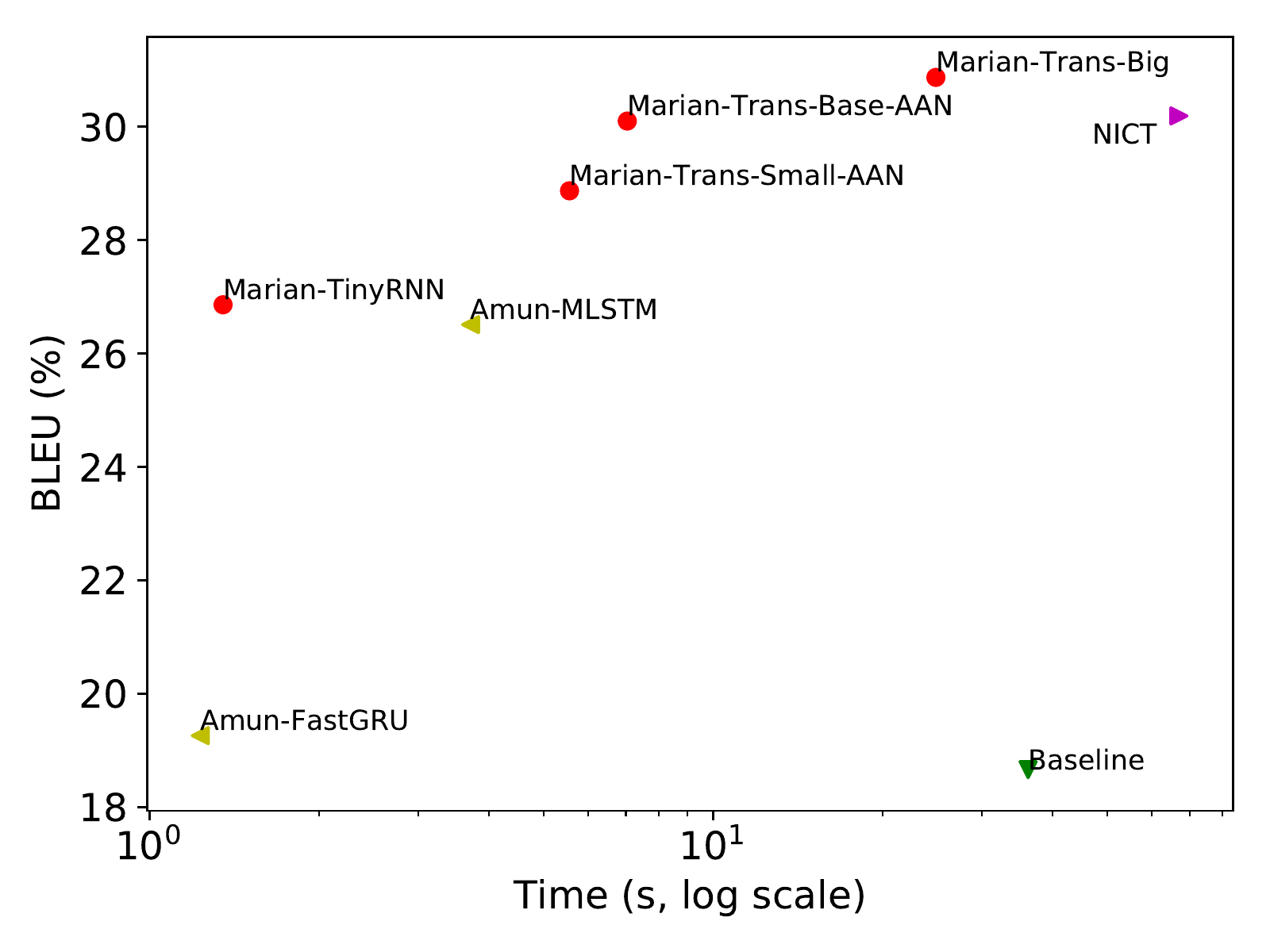} } \\
  \subfloat[CPU Memory vs. Accuracy]{\includegraphics[width=0.49\textwidth]{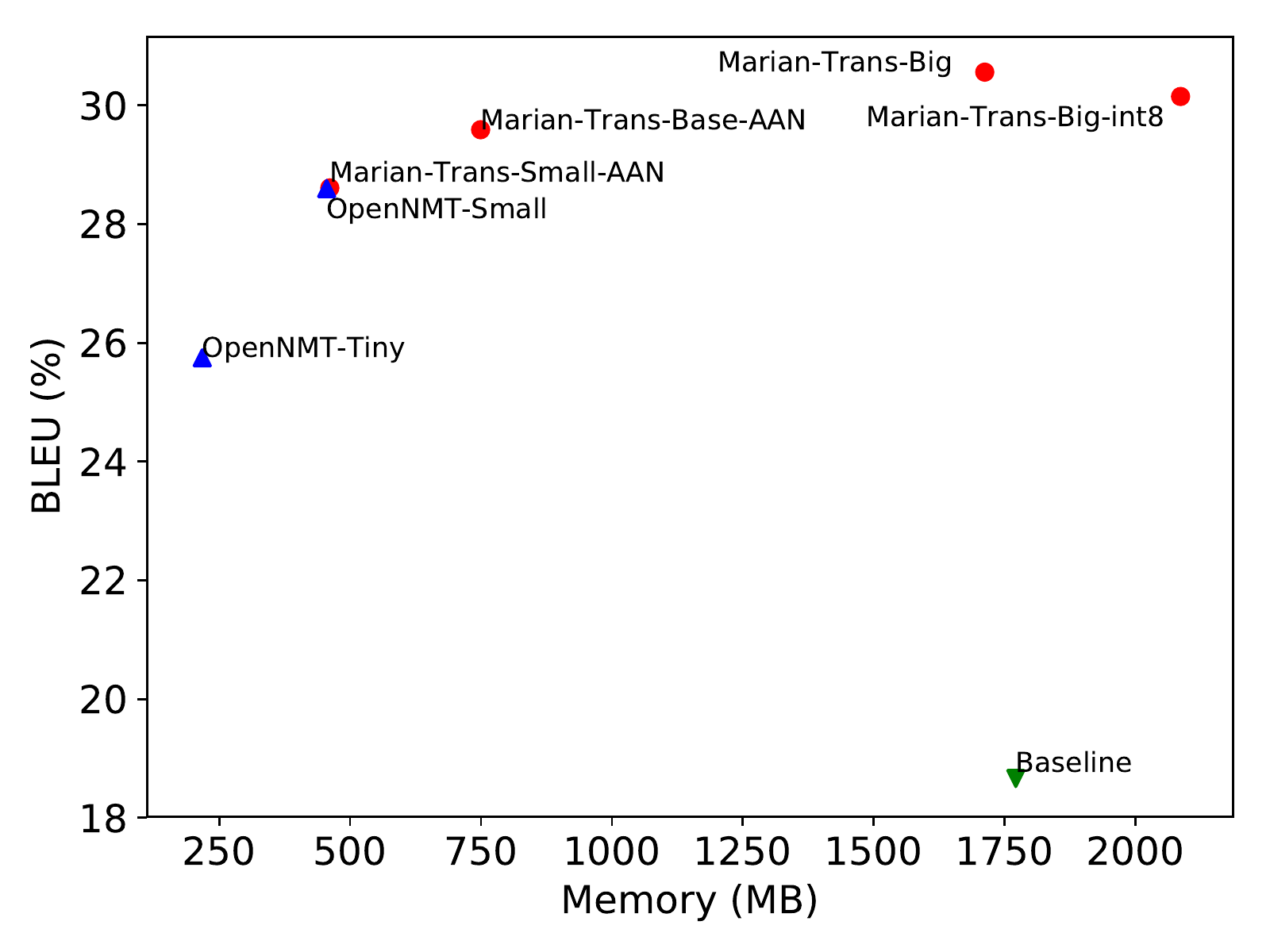} }
  \subfloat[GPU Memory vs. Accuracy]{\includegraphics[width=0.49\textwidth]{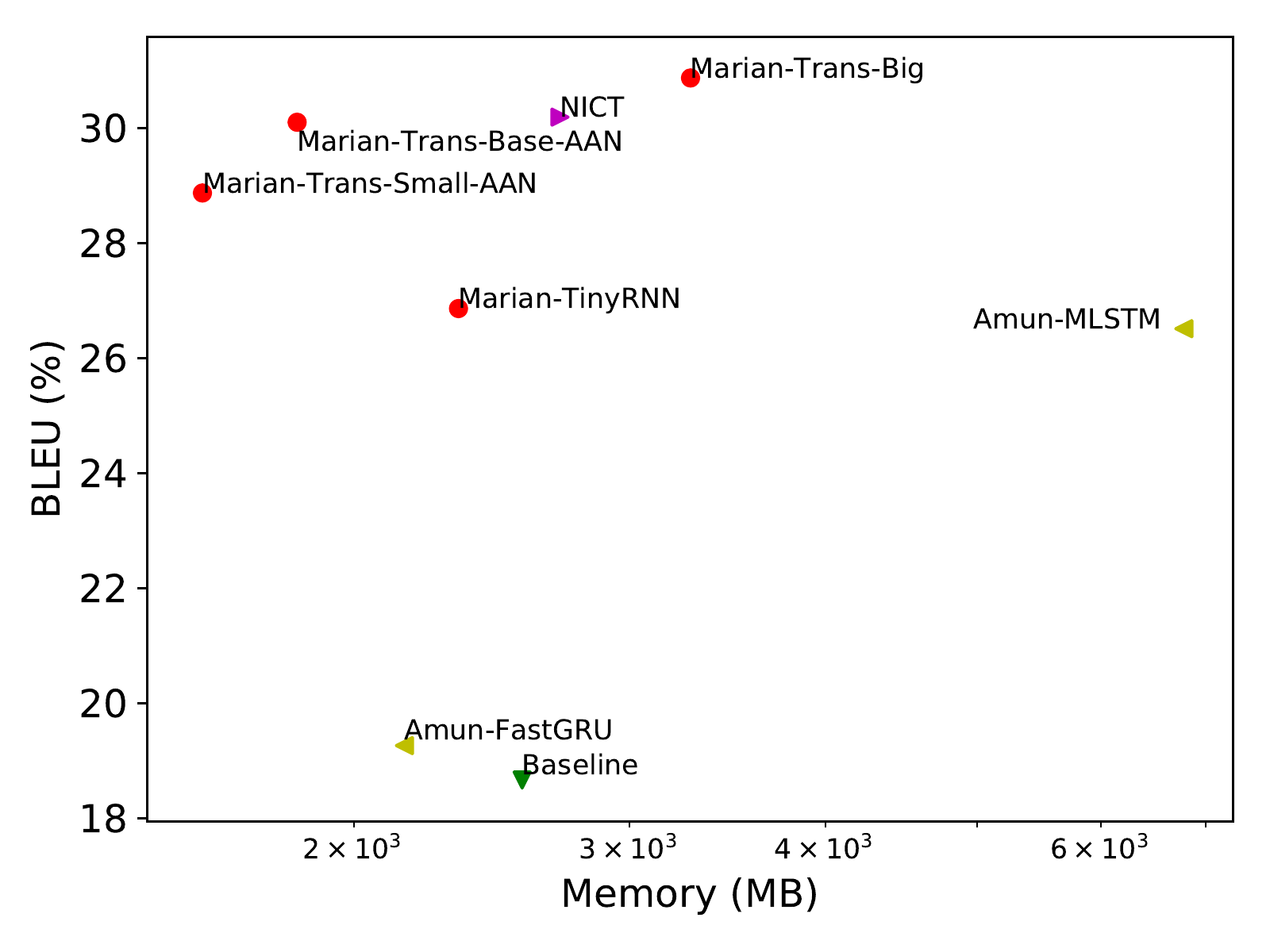} }
  \caption{ Time and memory vs. accuracy measured by BLEU, calculated on both CPU and GPU }
  \label{fig:results}
\end{figure*}

\subsection{Baseline Systems}

Two baseline systems were prepared:
\begin{description}
\item[Echo:] Just send the input back to the output.
\item[Base:] A baseline system using attentional LSTM-based encoder-decoders with attention \cite{bahdanau15alignandtranslate}.
\end{description}

\subsection{Submitted Systems}

Four teams, Team Amun, Team Marian, Team OpenNMT, and Team NICT submitted to the shared task, and we will summarize each below.
Before stepping in to the details of each system, we first note general trends that all or many systems attempted.
The first general trend was a fast C++ decoder, with Teams Amun, Marian, and NICT using the Amun or Marian decoders included in the Marian toolkit,\footnote{\url{https://marian-nmt.github.io}} and team OpenNMT using the C++-decoder decoder for OpenNMT.\footnote{\url{http://opennmt.net}}.
The second trend was the use of data augmentation techniques allowing the systems to train on data other than the true references.
Teams Amun, Marian, and OpenNMT all performed model distillation \cite{kim2016sequence}, where a larger teacher model is used to train a smaller student model, while team NICT used back translation, training the model on sampled translations from the target to the source \cite{imamura18selftraining}.
Finally, a common optimization was the use of lower-precision arithmetic, where Teams Amun, Marian, and OpenNMT all used some variety of 16/8-bit or integer calculation, along with the corresponding optimized CPU or GPU operations.
These three improvements seem to be best practices for efficient NMT implementation.

\subsubsection{Team Amun}

Team Amun's contribution \cite{hoang18amunwnmt} was based on the ``Amun'' decoder and consisted of a number of optimizations to improve translation speed on GPU.
The first major unique contribution was a strategy of batching together computations from multiple hypotheses within beam search to exploit parallelism of hardware.
Another contribution was a methodology to create a fused GPU kernel for the softmax calculation, that calculates all of the operations within the softmax (e.g. max, exponentiation, and sum) in a single kernel.
In the end they submitted two systems, \textbf{Amun-FastGRU} and \textbf{Amun-MLSTM}, which use GRU \cite{cho14phraserepresentations} and multiplicative LSTM \cite{krause2016multiplicative} units respectively.

\subsubsection{Team Marian}

Team Marian's system \cite{junczysdowmunt18marianwnmt} used the Marian C++ decoder, and concentrated on new optimizations for the CPU.
The team distilled a large self-attentional model into two types of ``student'' models: a smaller self-attentional model using average attention networks \cite{zhang2018accelerating}, a new higher-speed version of the original Transformer model \cite{vaswani2017attention}, and a standard RNN-based decoder.
They also introduced an auto-tuning approach that chooses which of multiple matrix multiplication implementations is most efficient in the current context, then uses this implementation going forward.
This resulted in the \textbf{Marian-TinyRNN} system using an RNN-based model, and the \textbf{Marian-Trans-Small-AAN}, \textbf{Marian-Trans-Base-AAN}, \textbf{Marian-Trans-Big}, \textbf{Marian-Trans-Big-int8} systems, which use different varieties and sizes of self-attentional models.

\subsubsection{Team OpenNMT}

Team OpenNMT \cite{senellart18opennmtwnmt} built a system based on the OpenNMT toolkit.
The model was based on a large self-attentional teacher model distilled into a smaller, fast RNN-based model.
The system also used a version of vocabulary selection \cite{shi17vocabulary}, and a method to increase the size of the encoder but decrease the size of the decoder to improve the efficiency of beam search. 
They submitted two systems, \textbf{OpenNMT-Small} and \textbf{OpenNMT-Tiny}, which were two variously-sized implementations of this model.

\subsubsection{Team NICT}

Team NICT's contribution \cite{imamura18nictwnmt} to the shared task was centered around using self-training as a way to improve NMT accuracy without changing the architecture.
Specifically, they used a method of randomly sampling pseudo-source sentences from a back-translation model \cite{imamura18selftraining} and used this to augment the data set to increase coverage.
They tested two basic architectures for the actual translation model, a recurrent neural network-based model trained using OpenNMT, and a self-attentional model trained using Marian, finally submitting the self-attentional model using Marian as their sole contribution to the shared task \textbf{NICT}.

\subsection{Shared Task Results}
\label{sec:sharedtaskresults}

A brief summary of the results of the shared task (for newstest2015) can be found in Figure \ref{fig:results}, while full results tables for all of the systems can be found in Appendix \ref{sec:fullresults}.
From this figure we can glean a number of observations.

First, encouragingly all the submitted systems handily beat the baseline system in speed and accuracy.

Secondly, observing the speed/accuracy curves, we can see that Team Marian's submissions tended to carve out the Pareto frontier, indicating that the large number of optimizations that went into creating the system paid off in aggregate.
Interestingly, on GPU, RNN-based systems carved out the faster but less accurate part of the Pareto curve, while on CPU self-attentional models were largely found to be more effective.
None of the submissions consisted of a Transformer-style model so small that it under-performed the RNN models, but a further examination of where the curves cross (if they do) would be an interesting examination for future shared tasks.

Next, considering memory usage, we can see again that the submissions from the Marian team tend to be the most efficient.
One exception is the extremely small memory system OpenNMT-Tiny, which achieves significantly lower translation accuracies, but fits in a mere 220MB of memory on the CPU.

In this first iteration of the task, we attempted to establish best practices and strong baselines upon which to build efficient test-time methods for NMT.
One characteristic of the first iteration of the task was that the basic model architectures used relatively standard, with the valuable contributions lying in solid engineering work and best practices in neural network optimization such as low-precision calculation and model distillation.
With these contributions, we now believe we have very strong baselines upon which future iterations of the task can build, examining novel architectures or methods for further optimizing the training speed.
We also will examine other considerations, such as efficient adaptation to new training data, or latency from receiving a sentence to translating it.

\section{Conclusion}

This paper summarized the results of the Second Workshop on Neural Machine Translation and Generation, where we saw a number of research advances, particularly in the area of efficiency in neural MT through submissions to the shared task.
The workshop series will continue next year, and continue to push forward the state of the art on these topics for faster, more accurate, more flexible, and more widely applicable neural MT and generation systems.

\section*{Acknowledgments}

We would like to warmly thank Amazon for its support of the shared task, both through its donation of computation time, and through its provision of a baseline system for participants to build upon. We also thank Google and Apple for their monetary support of the workshop.

\bibliography{myabbrv,acl2018}
\bibliographystyle{acl_natbib}

\appendix
\section{Full Shared Task Results}
\label{sec:fullresults}

For completeness, in this section we add tables of the full shared task results. These include the full size of the image file for the translation system (Table \ref{tab:image_file_sizes}), the comparison between compute time and evaluation scores on CPU (Table \ref{tab:time_and_eval_cpu}) and GPU (Table \ref{tab:time_and_eval_gpu}), and the comparison between memory and evaluation scores on CPU (Table \ref{tab:memory_cpu}) and GPU (Table \ref{tab:memory_gpu}).

\begin{table*}[t]
\begin{center}
\caption{Image file sizes of submitted systems.}
\label{tab:image_file_sizes}
\begin{tabular}{|ll|r|}
\hline
Team & System & Size [MiB] \\
\hline
edin-amun & fastgru                   & 4823.43 \\
          & mlstm.1280                & 5220.72 \\
Marian    & cpu-transformer-base-aan  &  493.20 \\
          & cpu-transformer-big       & 1085.93 \\
          & cpu-transformer-big-int8  & 1085.92 \\
          & cpu-transformer-small-aan &  367.92 \\
          & gpu-amun-tinyrnn          &  399.08 \\
          & gpu-transformer-base-aan  &  686.59 \\
          & gpu-transformer-big       & 1279.32 \\
          & gpu-transformer-small-aan &  564.32 \\
NICT      & marian-st                 & 2987.57 \\
OpenNMT   & cpu1                      &  339.02 \\
          & cpu2                      &  203.89 \\
Organizer & echo                      &  110.42 \\
          & nmt-1cpu                  & 1668.25 \\
          & nmt-1gpu                  & 3729.40 \\
\hline
\end{tabular}
\end{center}
\end{table*}
\begin{table*}[t]
\begin{center}
\caption{Time consumption and MT evaluation metrics (CPU systems).}
\label{tab:time_and_eval_cpu}
\resizebox{\textwidth}{!}{%
\begin{tabular}{|l|ll|r|rr|rr|}
\hline
\multirow{2}{*}{Dataset} &
\multirow{2}{*}{Team} &
\multirow{2}{*}{System} &
\multicolumn{3}{|c|}{Time Consumption [s]} &
\multirow{2}{*}{BLEU \%} &
\multirow{2}{*}{NIST} \\
& & & CPU & Real & Diff & & \\
\hline
Empty
  & Marian
    & cpu-transformer-base-aan  &    6.48 &    6.55 &     --- &   --- &   --- \\
  & & cpu-transformer-big       &    7.01 &    9.02 &     --- &   --- &   --- \\
  & & cpu-transformer-big-int8  &    7.31 &    7.51 &     --- &   --- &   --- \\
  & & cpu-transformer-small-aan &    6.32 &    6.33 &     --- &   --- &   --- \\
  & OpenNMT
    & cpu1                      &    0.64 &    0.65 &     --- &   --- &   --- \\
  & & cpu2                      &    0.56 &    0.56 &     --- &   --- &   --- \\
  & Organizer
    & echo                      &    0.05 &    0.06 &     --- &   --- &   --- \\
  & & nmt-1cpu                  &    1.50 &    1.50 &     --- &   --- &   --- \\
\hline
newstest2014
  & Marian
    & cpu-transformer-base-aan  &  281.72 &  281.80 &  275.25 & 27.44 & 7.362 \\
  & & cpu-transformer-big       & 1539.34 & 1541.00 & 1531.98 & 28.12 & 7.436 \\
  & & cpu-transformer-big-int8  & 1173.32 & 1173.41 & 1165.90 & 27.50 & 7.355 \\
  & & cpu-transformer-small-aan &  100.36 &  100.42 &   94.08 & 25.99 & 7.169 \\
  & OpenNMT
    & cpu1                      &  471.41 &  471.43 &  470.78 & 25.77 & 7.140 \\
  & & cpu2                      &   77.41 &   77.42 &   76.86 & 23.11 & 6.760 \\
  & Organizer
    & echo                      &    0.05 &    0.06 &    0.00 &  2.79 & 1.479 \\
  & & nmt-1cpu                  & 4436.08 & 4436.27 & 4434.77 & 16.79 & 5.545 \\
\hline
newstest2015
  & Marian
    & cpu-transformer-base-aan  &  223.86 &  223.96 &  217.41 & 29.59 & 7.452 \\
  & & cpu-transformer-big       & 1189.04 & 1190.97 & 1181.95 & 30.56 & 7.577 \\
  & & cpu-transformer-big-int8  &  907.95 &  908.43 &  900.92 & 30.15 & 7.514 \\
  & & cpu-transformer-small-aan &   80.21 &   80.25 &   73.92 & 28.61 & 7.312 \\
  & OpenNMT
    & cpu1                      &  368.93 &  368.95 &  368.30 & 28.60 & 7.346 \\
  & & cpu2                      &   59.02 &   59.02 &   58.46 & 25.75 & 6.947 \\
  & Organizer
    & echo                      &    0.05 &    0.06 &    0.00 &  3.24 & 1.599 \\
  & & nmt-1cpu                  & 3401.99 & 3402.14 & 3400.64 & 18.66 & 5.758 \\
\hline
\end{tabular}
}
\end{center}
\end{table*}

\begin{table*}[t]
\begin{center}
\caption{Time consumption and MT evaluation metrics (GPU systems).}
\label{tab:time_and_eval_gpu}
\resizebox{\textwidth}{!}{%
\begin{tabular}{|l|ll|r|rr|rr|}
\hline
\multirow{2}{*}{Dataset} &
\multirow{2}{*}{Team} &
\multirow{2}{*}{System} &
\multicolumn{3}{|c|}{Time Consumption [s]} &
\multirow{2}{*}{BLEU \%} &
\multirow{2}{*}{NIST} \\
& & & CPU & Real & Diff & & \\
\hline
Empty
  & edin-amun
    & fastgru                   &    4.18 &    4.24 &     --- &   --- &   --- \\
  & & mlstm.1280                &    4.44 &    4.50 &     --- &   --- &   --- \\
  & Marian
    & gpu-amun-tinyrnn          &    4.27 &    4.33 &     --- &   --- &   --- \\
  & & gpu-transformer-base-aan  &    5.62 &    5.68 &     --- &   --- &   --- \\
  & & gpu-transformer-big       &    6.00 &    6.05 &     --- &   --- &   --- \\
  & & gpu-transformer-small-aan &    5.48 &    5.54 &     --- &   --- &   --- \\
  & NICT
    & marian-st                 &    5.78 &    5.84 &     --- &   --- &   --- \\
  & Organizer
    & nmt-1gpu                  &    3.73 &    3.80 &     --- &   --- &   --- \\
\hline
newstest2014
  & edin-amun
    & fastgru                   &    5.68 &    5.74 &    1.50 & 17.74 & 5.783 \\
  & & mlstm.1280                &    8.64 &    8.70 &    4.20 & 23.85 & 6.833 \\
  & Marian
    & gpu-amun-tinyrnn          &    5.90 &    5.96 &    1.63 & 24.06 & 6.879 \\
  & & gpu-transformer-base-aan  &   14.58 &   14.64 &    8.95 & 27.80 & 7.415 \\
  & & gpu-transformer-big       &   36.74 &   36.80 &   30.74 & 28.34 & 7.486 \\
  & & gpu-transformer-small-aan &   12.46 &   12.52 &    6.97 & 26.34 & 7.219 \\
  & NICT
    & marian-st                 &   82.07 &   82.14 &   76.30 & 27.59 & 7.375 \\
  & Organizer
    & nmt-1gpu                  &   51.24 &   82.14 &   47.50 & 16.79 & 5.545 \\
\hline
newstest2015
  & edin-amun
    & fastgru                   &    5.41 &    5.47 &    1.23 & 19.26 & 5.905 \\
  & & mlstm.1280                &    8.15 &    8.22 &    3.71 & 26.51 & 7.015 \\
  & Marian
    & gpu-amun-tinyrnn          &    5.62 &    5.68 &    1.35 & 26.86 & 7.065 \\
  & & gpu-transformer-base-aan  &   12.67 &   12.73 &    7.04 & 30.10 & 7.526 \\
  & & gpu-transformer-big       &   30.81 &   30.90 &   24.84 & 30.87 & 7.630 \\
  & & gpu-transformer-small-aan &   11.04 &   11.10 &    5.56 & 28.87 & 7.379 \\
  & NICT
    & marian-st                 &   72.84 &   72.90 &   67.06 & 30.19 & 7.541 \\
  & Organizer
    & nmt-1gpu                  &   39.95 &   40.01 &   36.21 & 18.66 & 5.758 \\
\hline
\end{tabular}
}
\end{center}
\end{table*}

\begin{table*}[t]
\begin{center}
\caption{Peak memory consumption (CPU systems).}
\label{tab:memory_cpu}
\begin{tabular}{|l|ll|rrr|}
\hline
\multirow{2}{*}{Dataset} &
\multirow{2}{*}{Team} &
\multirow{2}{*}{System} &
\multicolumn{3}{|c|}{Memory [MiB]} \\
& & & Host & GPU & Both \\
\hline
Empty
  & Marian
    & cpu-transformer-base-aan  &  531.39 & --- &  531.39 \\
  & & cpu-transformer-big       & 1768.56 & --- & 1768.56 \\
  & & cpu-transformer-big-int8  & 1193.13 & --- & 1193.13 \\
  & & cpu-transformer-small-aan &  367.22 & --- &  367.22 \\
  & OpenNMT
    & cpu1                      &  403.86 & --- &  403.86 \\
  & & cpu2                      &  194.61 & --- &  194.61 \\
  & Organizer
    & echo                      &    1.15 & --- &    1.15 \\
  & & nmt-1cpu                  & 1699.71 & --- & 1699.71 \\
\hline
newstest2014
  & Marian
    & cpu-transformer-base-aan  &  761.07 & --- &  531.39 \\
  & & cpu-transformer-big       & 1681.81 & --- & 1768.56 \\
  & & cpu-transformer-big-int8  & 2084.66 & --- & 1193.13 \\
  & & cpu-transformer-small-aan &  476.21 & --- &  367.22 \\
  & OpenNMT
    & cpu1                      &  458.08 & --- &  403.86 \\
  & & cpu2                      &  219.79 & --- &  194.61 \\
  & Organizer
    & echo                      &    1.20 & --- &    1.15 \\
  & & nmt-1cpu                  & 1770.69 & --- & 1699.71 \\
\hline
newstest2015
  & Marian
    & cpu-transformer-base-aan  &  749.29 & --- &  531.39 \\
  & & cpu-transformer-big       & 1712.08 & --- & 1768.56 \\
  & & cpu-transformer-big-int8  & 2086.02 & --- & 1193.13 \\
  & & cpu-transformer-small-aan &  461.27 & --- &  367.22 \\
  & OpenNMT
    & cpu1                      &  455.21 & --- &  403.86 \\
  & & cpu2                      &  217.64 & --- &  194.61 \\
  & Organizer
    & echo                      &    1.11 & --- &    1.15 \\
  & & nmt-1cpu                  & 1771.35 & --- & 1699.71 \\
\hline
\end{tabular}
\end{center}
\end{table*}

\begin{table*}[t]
\begin{center}
\caption{Peak memory consumption (GPU systems).}
\label{tab:memory_gpu}
\begin{tabular}{|l|ll|rrr|}
\hline
\multirow{2}{*}{Dataset} &
\multirow{2}{*}{Team} &
\multirow{2}{*}{System} &
\multicolumn{3}{|c|}{Memory [MiB]} \\
& & & Host & GPU & Both \\
\hline
Empty
  & edin-amun
    & fastgru                   &  442.55 &  668 & 1110.55 \\
  & & mlstm.1280                &  664.88 &  540 & 1204.88 \\
  & Marian
    & gpu-amun-tinyrnn          &  346.80 &  522 &  868.80 \\
  & & gpu-transformer-base-aan  &  487.18 &  484 &  971.18 \\
  & & gpu-transformer-big       & 1085.29 &  484 & 1569.29 \\
  & & gpu-transformer-small-aan &  366.26 &  484 &  850.26 \\
  & NICT
    & marian-st                 &  510.43 &  484 &  994.43 \\
  & Organizer
    & nmt-1gpu                  &  378.81 &  640 & 1018.81 \\
\hline
newstest2014
  & edin-amun
    & fastgru                   &  456.29 & 1232 & 1688.29 \\
  & & mlstm.1280                &  686.29 & 5144 & 5830.29 \\
  & Marian
    & gpu-amun-tinyrnn          &  346.93 & 1526 & 1872.93 \\
  & & gpu-transformer-base-aan  &  492.91 & 1350 & 1842.91 \\
  & & gpu-transformer-big       & 1081.69 & 2070 & 3151.69 \\
  & & gpu-transformer-small-aan &  366.55 & 1228 & 1594.55 \\
  & NICT
    & marian-st                 &  922.53 & 1780 & 2702.53 \\
  & Organizer
    & nmt-1gpu                  &  377.18 & 2178 & 2555.18 \\
\hline
newstest2015
  & edin-amun
    & fastgru                   &  473.72 & 1680 & 2153.72 \\
  & & mlstm.1280                &  684.84 & 6090 & 6774.84 \\
  & Marian
    & gpu-amun-tinyrnn          &  350.19 & 1982 & 2332.19 \\
  & & gpu-transformer-base-aan  &  489.62 & 1350 & 1839.62 \\
  & & gpu-transformer-big       & 1082.52 & 2198 & 3280.52 \\
  & & gpu-transformer-small-aan &  372.56 & 1228 & 1600.56 \\
  & NICT
    & marian-st                 &  929.70 & 1778 & 2707.70 \\
  & Organizer
    & nmt-1gpu                  &  383.02 & 2178 & 2561.02 \\
\hline
\end{tabular}
\end{center}
\end{table*}

\end{document}